\documentclass[11pt,a4paper]{article}
\PassOptionsToPackage{hyphens}{url}
\usepackage[hyperref]{eacl2021}
\usepackage{times}
\usepackage{latexsym}
\usepackage{booktabs} %
\usepackage{graphicx}
\usepackage{multicol}
\usepackage{multirow}
\usepackage{xspace}
\usepackage{amsmath}

\usepackage{microtype}

\DeclareMathOperator*{\argmax}{arg\,max}

\aclfinalcopy %

\newcommand\name{LOME\xspace}

\newcommand{\jhu}{\textsuperscript{\rm{1}}}
\newcommand{\urochester}{\textsuperscript{\rm{2}}}

\title{\name: Large Ontology Multilingual Extraction}

\author{Patrick Xia\jhu\thanks{~~Equal Contribution}~, Guanghui Qin\jhu\footnotemark[1]~, Siddharth Vashishtha\urochester \\
  \textbf{Yunmo Chen\jhu, Tongfei Chen\jhu, Chandler May\jhu, Craig Harman\jhu} \\
  \textbf{Kyle Rawlins\jhu, Aaron Steven White\urochester, Benjamin Van Durme\jhu} \\
  \jhu~Johns Hopkins University, \urochester~University of Rochester\\
  \texttt{\string{paxia,qin,vandurme\string}@jhu.edu}}

\date{}

\begin{document}
\maketitle
\begin{abstract}
We present \name, a system for performing multilingual information extraction. Given a text document as input, our core system identifies spans of textual entity and event mentions with a FrameNet \citep{baker-etal-1998-berkeley} parser. It subsequently performs coreference resolution, fine-grained entity typing, and temporal relation prediction between events. By doing so, the system constructs an event and entity focused knowledge graph. We can further apply third-party modules for other types of annotation, like relation extraction. Our (multilingual) first-party modules either outperform or are competitive with the (monolingual) state-of-the-art. We achieve this through the use of multilingual encoders like \textit{XLM-R} \cite{conneau-etal-2020-unsupervised} and leveraging multilingual training data. \name is available as a Docker container on Docker Hub. In addition, a lightweight version of the system is accessible as a web demo.

\end{abstract}

\section{Introduction}

\begin{figure*}[t]
    \centering
    \includegraphics[width=\textwidth]{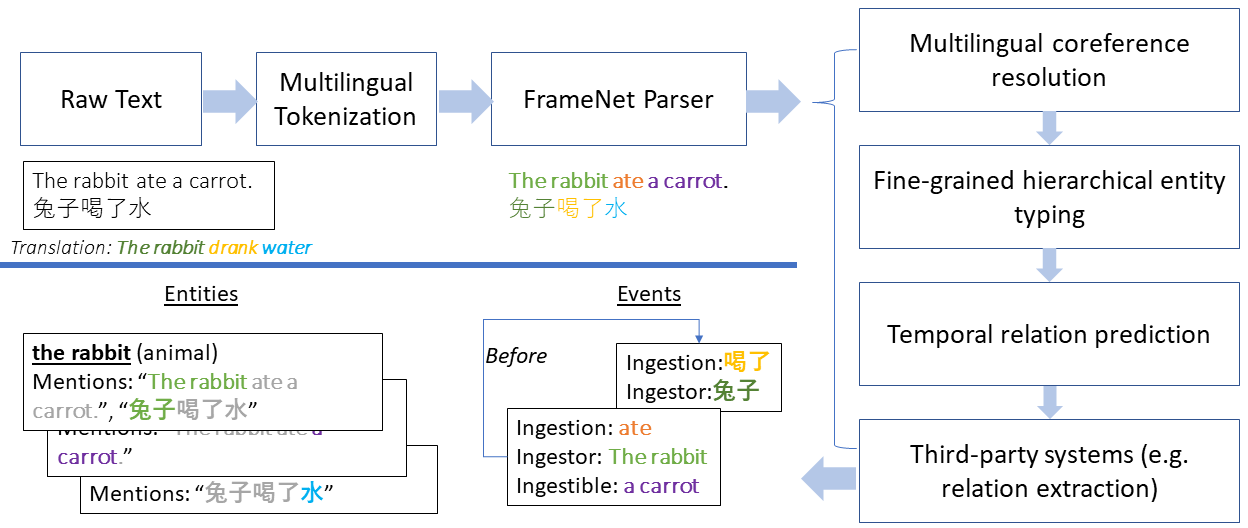}
    \caption{Architecture of \name. The system processes text documents as input and first uses a FrameNet parser to detect entities and events. Then, a suite of models enrich the entities and events with additional predictions. Each individual model can be trained and tuned independently, ensuring modularity of the pipeline. Annotations between models are transferred using \textsc{Concrete}, a data schema for NLP.}
    \label{fig:lome_architecture}
\end{figure*}

As information extraction capabilities continue to improve due to advances in modeling, encoders, and data collection, we can now look (back) toward making richer predictions at the document-level, with a large ontology, and across multiple languages. Recently, \citet{li-etal-2020-gaia} noted that despite a growth of open-source NLP software in general, there is still a lack of available software for knowledge extraction. We wish to provide a starting point that allows others to build increasingly comprehensive document-level knowledge graphs of events and entities from text in many languages.\footnote{Information on using the Docker container, web demo, and demo video at \url{https://nlp.jhu.edu/demos}.}

Therefore, we demonstrate LOME, a system for multilingual information extraction with large ontologies. \autoref{fig:lome_architecture} shows the high-level pipeline by following a multilingual input example. A sentence-level parser identifies both \textsc{ingestion} events and their arguments. To connect these events cross-sententially, the system clusters coreferent mentions and predicts the temporal relations between the events. LOME, which supports fine-grained entity types, additionally labels entities like \underline{\textbf{the rabbit}} with \textsc{living\_thing/animal}.  

Several prior packages have also used advances in state-of-the-art models to build comprehensive information extraction systems. \citet{li-etal-2019-multilingual} present an event, relation, and entity extraction and coreference system for three languages: English, Russian, and Ukrainian. \citet[GAIA]{li-etal-2020-gaia} extend that work to support cross-media documents. However, both of these systems consist of language-specific models that operate on monolingual documents after first identifying the language. On the other hand, work prioritizing coverage across tens or hundreds of languages is limited in their scope in extraction \cite{akbik-li-2016-polyglot, pan-etal-2017-cross}.

Like prior work, \name is focused on extracting entities and events from raw text documents. However, \name is language-agnostic; all components prioritize multilinguality. Using \textit{XLM-R} \cite{conneau-etal-2020-unsupervised} as the underlying encoder paves the way for both training on multilingual data (where it exists) and inference in many languages.\footnote{\textit{XLM-R} itself is trained on CommonCrawl data spanning one hundred languages.} Our pipeline includes a full FrameNet parser for events and their arguments, neural coreference resolution, an entity typing model over large ontologies, and temporal resolution between events. 

Our system is designed to be modular: each component is trained independently and tuned on task-specific data. To communicate between modules, we use \textsc{Concrete} \cite{ferraro2014concretely}, a data schema used in other text processing systems \cite{peng-etal-2015-concrete}. One advantage of using a standardized data schema is that it enables modularization and extension. Unless there are annotation dependencies, individual modules can be inserted, replaced, merged, or bypassed depending on the application. We discuss two example applications of our \textsc{Concrete}-based modules, one of which further extracts relations and the other performs cross-sentence argument linking for events. 

\section{Tasks}

The overarching application of \name is to extract an entity- and event-centric knowledge graph from a textual document. In particular, we are interested in using these graphs to support a multilingual schema learning task (KAIROS\footnote{This goal is to develop a system that identifies, links, and temporally sequences complex events. More information at \url{https://www.darpa.mil/program/knowledge-directed-artificial-intelligence-reasoning-over-schemas}.}) for which data has been annotated by the LDC \cite{cieri-etal-2020-progress}. As a result, some parts of LOME are designed for compatibility with the KAIROS event and entity ontology. Nonetheless, there is significant overlap with publicly available datasets, which we describe for those tasks.

\autoref{fig:lome_architecture} presents the architecture of our pipeline. Besides the FrameNet parser, which is run first, the remaining modules can be run in any order, if at all. In addition, our use of a standardized data schema for communication allows for the integration of third-party systems. In this section, we will go into further detail for each task.

\subsection{FrameNet Parsing}

FrameNet parsing is a semantic role labeling style task. The goal is to find all the frames and their roles, as well as the trigger spans associated with them in a sentence. Frames are concepts, such as events or entities, in a sentences.
Every frame is associated with some roles, and both of them are triggered by spans in the sentence.

Unlike most previous work \citep{yang-mitchell-2017-joint,peng-etal-2018-learning,swayamdipta-etal-2018-syntactic}, our system is not conditioned on the trigger spans or frames. We perform ``full parsing'' \citep{das-etal-2014-frame}, where the input is a raw sentence, and the output is the complete structure predictions.

As the first model in the whole pipeline system, the trigger spans found by the FrameNet parser will be used as candidate spans for all other tasks.

\subsection{Entity Coreference Resolution}

In coreference resolution, the goal is to cluster spans in the text that refer to the same entity. Neural models for doing so typically encode the text first before identifying possible mentions \cite{lee-etal-2017-end, joshi-etal-2019-bert, joshi-etal-2020-spanbert}. These spans are scored pairwise to determine whether two spans refer to each other. These scores then determine coreference clusters by decoding under a variety of strategies \cite{lee-etal-2018-higher, xu-choi-2020-revealing}.

In this work, we choose a constant-memory variant of that model which also achieves high performance \cite{xia-etal-2020-incremental}. The motivation here is robustness: we prioritize the ability to soundly run on all document lengths over slightly better performing but fragile systems. In addition, because this coreference resolution model is part of a broader entity-centric system, the module used in this system does not perform the mention detection step (which is left to the FrameNet parser). Instead, both training and inference assumes given mentions, and the task we are concerned about in this paper is mention \textit{linking}.

\subsection{Entity Typing}

\begin{figure}[t]
    \centering
    \includegraphics[width=0.67\linewidth]{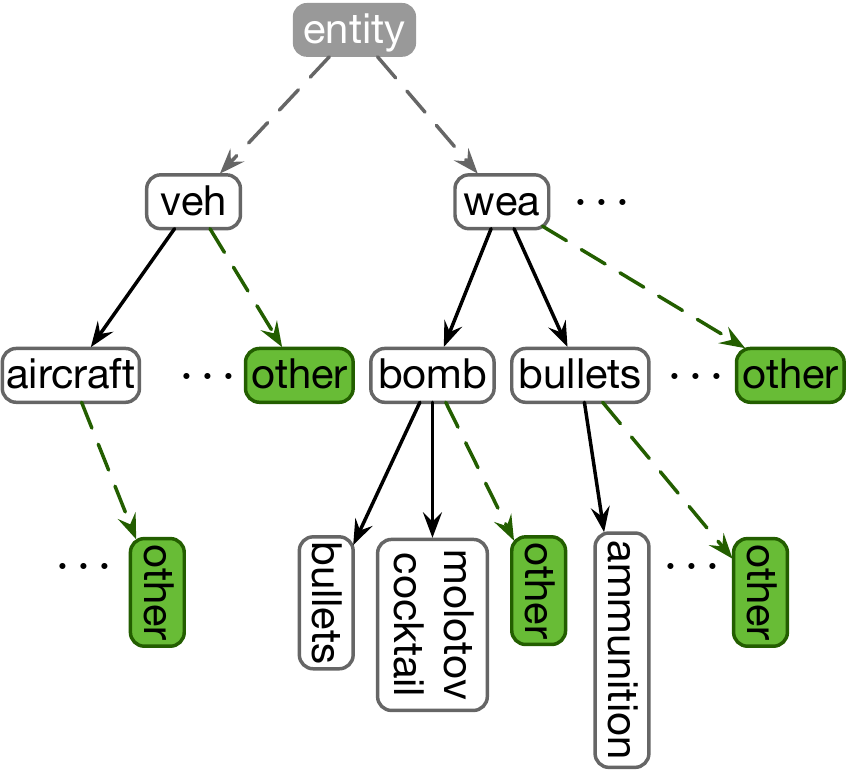}
    \caption{A portion of the AIDA entity type ontology.}
    \label{fig:ontology-aida}
\end{figure}

Entity typing assigns a fine-grained semantic label to a span of text, where the span is a \emph{mention} of some entity found by the FrameNet parser. Traditionally, labels include \texttt{PER}, \texttt{GPE}, \texttt{ORG}, etc., but recent work in \textit{fine-grained} entity typing seek to classify spans into types defined by hierarchical type \textit{ontologies} (e.g. BBN \cite{WeischedelB05}, FIGER \cite{LingW12}, UltraFine\footnote{UltraFine is slightly different in that the types are bucketed into 3 categories of different granularity, but without explicit subtyping relations.} \cite{choi-etal-2018-ultra}, COLLIE \citep{allen-etal-2020-broad}). Such ontologies refine coarse types like \texttt{PER} to fine-grained types such as \texttt{/person/artist/singer} that sits on a type hierarchy. A portion of the AIDA ontology (LDC2019E07) is illustrated in \autoref{fig:ontology-aida}.

To support fine-grained ontologies, we employ a recent coarse-to-fine-decoding entity typing model \cite{chen-etal-2020-hierarchical} that is specifically designed to assign types that are defined by hierarchical ontologies. The use of a coarse-to-fine model also allows users to select between coarse- and fine-grained types. We swap the underlying encoder from ELMo \cite{peters-etal-2018-deep} to \textit{XLM-R} to be able to assign types over mentions in different languages using a single multilingual model, and to enable transfer between languages.

The base typing model in \citet{chen-etal-2020-hierarchical} supports entity typing on entity \emph{mentions}. We extend this model to gain the ability to perform entity typing on \emph{entities}, i.e. clusters of entity mentions. Since our decoder is coarse-to-fine and predicts a type at each level of the type hierarchy, we employ Borda voting on each level. Specifically, given a coreference chain comprising mentions $m_{1, \cdots, n}$, and the score for mention $m_i$ being typed as type $t$ as $s_{i,t}$, we perform Borda counting to select the most confident type $t^* = \argmax_t \sum_i r(i,t)$ over all $t$'s in a specific type level, where $r(i,t) = 1/\mathrm{rank}_t(s_{i,t})$ is the ranking relevance score used in Borda counting.

\subsection{Temporal Relation Extraction}

The task of temporal relation extraction focuses on finding the chronology of events (e.g., \textit{Before}, \textit{After}, \textit{Overlaps}) in text. Extracting temporal relation is useful for various downstream tasks -- curating structured clinical data \citep{savova2010mayo,soysal2018clamp}, text summarization \citep{Glavas2014EventGF,kedzie-etal-2015-predicting}, question-answering \citep{llorens-etal-2015-semeval,zhou-etal-2019-going}, etc. The task is most commonly viewed as a classification task where given a pair of events and its textual context, the temporal relation between them needs to be identified. 

The construction of the TimeBank corpus \citep{pustejovsky2003timebank} largely spurred the research in temporal relation extraction. It included 14 temporal relation labels. Other corpora \citep{verhagen-etal-2007-semeval,verhagen-etal-2010-semeval,sun2013evaluating,cassidy-etal-2014-annotation} reduced the number of labels to a smaller number owing to lower inter-annotator agreements and sparse annotations. Various types of models \citep{chambers-etal-2014-dense,cheng-miyao-2017-classifying,leeuwenberg-moens-2017-structured,ning-etal-2017-structured,vashishtha-etal-2019-fine,zhou2020clinical} have been used in the recent years to extract temporal relations from text.

In this work, we use \citet{vashishtha-etal-2019-fine}'s best model and retrain it using \textit{XLM-R}.  We evaluate their model using the transfer learning approach described in their work and retrain it on TimeBank-Dense (TBD) \citep{cassidy-etal-2014-annotation}. TBD uses a reduced set of 5 temporal relation labels -- \textit{before}, \textit{after}, \textit{includes}, \textit{is\_included}, and \textit{vague}.

\section{System Design}

\subsection{Modularization}

Our system is modularized into separate models and libraries that communicate with each other using \textsc{Concrete}, a data format for richly annotating natural language documents \cite{ferraro2014concretely}. Each component is independent of each other, which allows for both inserting additional modules or deleting those provided in the default pipeline. We choose this loosely-affiliated design to enable both faster and independent prototyping of individual components, as well as better compartmentalization of our models.

We emphasize that the system is a pipeline: while individual modules can be further improved, the system is not designed to be trained end-to-end and benchmarking the richly-annotated output depends on the application and priorities. In this paper, we only benchmark individual components and describe a couple of applications.

\subsection{System Inputs and Outputs}

The system can consume, as input, either tokenized or untokenized text, which is first tokenized either by whitespace or with a multilingual tokenizer, PolyGlot.\footnote{\url{https://github.com/aboSamoor/polyglot}} However, this tokenization is not necessarily used by all modules, which may choose to either operate on the raw text itself or on a SentencePiece \cite{kudo-richardson-2018-sentencepiece} retokenization. 

The system outputs a \textsc{Concrete} communication file for each input document. This output file contains annotations including entities, events, coreference, entity types, and temporal relations. This schema used is entirely self-contained and the well-documented library also contains tools for visualizing and inspecting \textsc{Concrete} files.\footnote{\url{http://hltcoe.github.io/concrete/}} For the web demo, the output is displayed in the browser. 

\section{Evaluation Benchmarks}

\subsection{FrameNet Span Finding}

The FrameNet parser is comprised of an \textit{XLM-R} encoder, a BIO tagger, and a typing module.
It encodes the input sentences into a list of vectors, used by both the BIO tagger and the typing module.
The goal of BIO tagger is to find trigger spans, which are then labeled by the typing module.
To parse a sentence, we run the model to find all frames, and then find their roles conditioned on the frames.

We train the FrameNet parser on the FrameNet v1.7 corpus following \citet{das-etal-2014-frame}, with statistics in \autoref{tab:fn_stats}.
We evaluate the results with exact matching as our metric,\footnote{A role is considered to be correctly predicted only when its frame is precisely predicted.}
and get 56.34 labeled F1 or 66.41 unlabeled F1.
Since we are not aware of previous work on both full parsing and a metric for its evaluation, we do not have a baseline.
However, we can force the model to perform frame identification given the trigger span, like prior work. These results are shown in \autoref{tab:frame_id}.

\begin{table}[htb!]
    \small
    \centering
    \begin{tabular}{lllll}
    \toprule
     & \# Sentences & \# Frames & \# Roles \\
    \midrule 
    train & 3120 & 18604 & 32419 \\
    dev & 311 & 2209 & 3853\\
    test & 1333 & 6687 & 11277 \\
    \bottomrule
    \end{tabular}
    \caption{Statistics of FrameNet v1.7}
    \label{tab:fn_stats}
\end{table}

\begin{table}[htb!]
    \small
    \centering
    \begin{tabular}{ll}
    \toprule
    \textbf{Model} & Accuracy \\
    \midrule \citet{yang-mitchell-2017-joint} & 88.2 \\
    \citet{hermann-etal-2014-semantic} & 88.4 \\
    \citet{peng-etal-2018-learning} & 90.0 \\
    This work & \textbf{91.3} \\
    \bottomrule
    \end{tabular}
    \caption{Result on frame identification}
    \label{tab:frame_id}
\end{table}

\subsection{Coreference Resolution}

We retrain the model by \citet{xia-etal-2020-incremental} with \textit{XLM-R} (large) as the underlying encoder and with additional multilingual data. The model is a constant-memory variant of neural coreference resolution models. We refer the reader to \citet{xia-etal-2020-incremental} for model and training details.

Unlike that work, we operate under the assumption that we are provided gold spans. This is motivated by the location of coreference in \name. In addition, while they use a frozen encoder, we found that finetuning improves performance.\footnote{We use AdamW and a learning rate of $5\times 10^{-6}$.} Finally, we train on the full OntoNotes 5.0 \cite{weischedel2013ontonotes, pradhan-etal-2013-towards}, a subset of SemEval 2010 Task 1 \cite{recasens-etal-2010-semeval}, and two additional sources of Russian data, RuCor \cite{azerkovich2014evaluating6887693} and AnCor \cite{Budnikov2019rueval}.

We benchmark the performance of our model on each language. We report the average F1 of MUC \cite{vilain-etal-1995-model}, B$^3$ \cite{Bagga98algorithmsfor}, and CEAF$_{\phi_4}$ \cite{luo-2005-coreference} by language in \autoref{tab:coref}. %
We can compare the model's performance to monolingual gold-only baselines, where they exist. For English, we trained an identical model but instead use SpanBERT \cite{joshi-etal-2020-spanbert}, an English-only encoder finetuned for English OntoNotes coreference. That model achieves 92.2 average (dev.) F1, compared to our 92.7. There is also a comparable system for Russian AnCor from \citet{le19sentence}, which achieves 79.9 F1 using the model from \citet{lee-etal-2018-higher} and RuBERT \cite{kuratov2019adaptation}. This shows that our single, multilingual model, can perform similarly to monolingual models, with the advantage that our model does not need to perform language ID. This finding mirrors prior findings showing multilingual encoders are strong cross-lingually \cite{wu-dredze-2019-beto}.

\begin{table}[h]
    \small
    \centering
    \begin{tabular}{rccccr}
    \toprule
     Language & \# Training & \# Eval Docs & Avg. F1  \\
    \midrule
     Arabic\textsuperscript{\textsc{o}} & 359 & 44 & 71.3 \\
     Catalan\textsuperscript{\textsc{s}} & 829 & 142 & 58.7 \\
     Chinese\textsuperscript{\textsc{o}} & 1810 & 252 &  90.8 \\
     Dutch\textsuperscript{\textsc{s}} & 145 & 23 &  63.5 \\
     English\textsuperscript{\textsc{o}} & 2802 & 343 & 92.7 \\
     Italian\textsuperscript{\textsc{s}} & 80 & 17 &  47.2\\
     Russian\textsuperscript{\textsc{a}} & 573 & 127 & 77.3 \\
     Spanish\textsuperscript{\textsc{s}} & 875 & 140 & 63.5\\
    \bottomrule
    \end{tabular}
    \caption{Average F1 scores by language with gold mentions. The superscripts \textsc{o} indicates data from OntoNotes 5.0 (dev), \textsc{s} indicates data from SemEval 2010 Task 1 (dev), and \textsc{a} is the AnCor data (test).}
    \label{tab:coref}
\end{table}

\subsection{Entity Typing}

We retrain the coarse-to-fine entity typer by \citet{chen-etal-2020-hierarchical} with \textit{XLM-R} as the underlying encoder, and using the AIDA ontology as the type label inventory.
The dataset annotated from AIDA is relatively small. To make the model more robust, we pre-train the model using extra training data from GAIA \citep{li-etal-2020-gaia}, where they obtained YAGO fine-grained types \cite{SuchanekKW08} from the results of Freebase entity linking, and mapped these types to the AIDA ontology.
After pre-training, we fine-tune the model using the AIDA M18 and M36 data with 3-fold cross-validation, where each fold is distinct in the topics of these documents. The sizes of these datasets are shown in \autoref{tab:entity-typing-stats}.

\begin{table}[h]
    \centering \small
    \begin{tabular}{ccc}
      \toprule
        \bf Data source & \bf Language & \bf \# of entities \\
      \midrule
      \multirow{2}{*}{AIDA M18} & English & 4,433 \\
                                & Russian & 4,826 \\
            \tt LDC2019E07      & Ukrainian & 4,261 \\
      \midrule
      \multirow{2}{*}{AIDA M36} & English & 703 \\
                                & Spanish & 557 \\
              \tt LDC2020E29    & Russian & 729 \\
      \midrule
      \multirow{3}{*}{GAIA}     & English & 42.8M \\
                                & Spanish & 11.1M \\
                                & Russian & 2.4M  \\
      \bottomrule
    \end{tabular}
    \caption{Statistics of the datasets used for training our entity typing model.}
    \label{tab:entity-typing-stats}
\end{table}

Our models perform well in these datasets. Using one third of the AIDA M36 data as dev, our method obtained 60.1\% micro-$\rm F_1$ score;\footnote{Please refer to \citet{chen-etal-2020-hierarchical} for the exact definitions of the evaluation metric.} with pre-training using GAIA extra data, we get 76.5\%.

Our system can also be extended to support other commonly used fine-grained entity type ontologies. We report the results in micro-$\rm F_1$ in \autoref{tab:typer}.
\begin{table}[h]
  \centering\small
  \begin{tabular}{cll}
  \toprule
    \bf Ontology & \bf Prior state-of-the-art & \bf Ours \\
  \midrule
    BBN   &     78.1 \cite{lin-ji-2019-attentive} & \bf 80.5 \\
    FIGER &     79.8 \cite{lin-ji-2019-attentive} & \bf 80.8 \\
    UltraFine & 40.1 \cite{onoe-durrett-2019-learning} & \bf 41.5 \\
  \bottomrule
  \end{tabular}
  \caption{Performance of our hierarchical entity typing model across several typing ontologies.}
  \label{tab:typer}
\end{table}

\subsection{Temporal Relation Extraction}

We retrain \citet{vashishtha-etal-2019-fine}'s best fine-grained temporal relation model on UDS-T \citep{vashishtha-etal-2019-fine} using \textit{XLM-R} (large). We then use their transfer learning approach and train an SVM model on event-event relations in TimeBank-Dense (TBD) to predict categorical temporal relation labels. With this approach, we see a micro-F1 score of 56 on the test set of TBD.\footnote{The train and dev set of TBD has a total of 4,590 instances and the test set has 1,405 instances of event-event relations.}

For better performance, we train the same model on additional TempEval3 (TE3) dataset \citep{uzzaman-etal-2013-semeval}. Since TE3 and TBD use a different set of temporal relations, we consider only those instances that are labeled with 4 temporal relations from both TE3 and TBD for joint training -- \textit{before}, \textit{after}, 
\textit{includes} (\textit{container}), and \textit{is\_included} (\textit{contained}). We retrain \citet{vashishtha-etal-2019-fine}'s transfer learning model on the combined TE3 and TBD dataset considering only these 4 relations and evaluate on their combined test set.\footnote{We consider only event-event relations and the combined dataset has 5,987 (1,249) instances in the train (test) set.} Results on the combined test set are reported in Table \ref{tab:temporal_eval}. We use this model as the default temporal relation extraction model in LOME.

We also test our default model on a Chinese temporal relation extraction dataset \citep{li-etal-2016-global}.\footnote{We remove the instances with \textit{unknown} relation from the dataset and convert the predictions with \textit{includes} and \textit{is\_included} relations to the \textit{overlaps} relation to match the label set of their dataset with our system.} In the zero-shot setting, we get a micro F1 score of 52.6 on the provided dataset, as compared to a majority baseline of 37.5.\footnote{The authors were able to provide only half of the dataset with 10,476 event-event pairs, from which we ignore instances with \textit{unknown} relation, resulting into 9,362 instances.} Similar to the default temporal system in LOME, we use the \textit{XLM-R} version of \citet{vashishtha-etal-2019-fine}'s model obtaining relation embeddings for the Chinese dataset and train an SVM model using the transfer learning approach to get a micro F1 score of 64.4.\footnote{The results are the average of the 5-fold cross validation splits provided by \citet{li-etal-2016-global}.}

\begin{table}[htb!]
    \small
    \centering
    \begin{tabular}{cccc}
    \toprule
    \textbf{Relation} & \textbf{Precision} & \textbf{Recall} & \textbf{\ F1}  \\
    \midrule
    before & 68 & 89 & 77\\
    after & 74 & 69 & 71\\
    includes & 83 & 5 & 10\\
   is\_included & 44 & 15 & 22\\
    \bottomrule
    \end{tabular}
    \caption{Result on the combined test set of TempEval3 and TimeBank-Dense when trained with just 4 temporal relation labels}
    \label{tab:temporal_eval}
\end{table}

\section{Extensions}

\subsection{Incorporating third-party systems}

Besides the core components described above, we also discuss the viability of including additional modules that may not fit directly in the core pipeline but can be included depending on the downstream application. For example, the system described above does not predict any relation information, which is needed for the motivating application of downstream schema inference. 
To do so, we wrote a \textsc{Concrete} and Docker wrapper around OneIE \cite{lin-etal-2020-joint} and attached it at the end of the pipeline. %
With our \textsc{Concrete} based design, the integration of any third-party module can be done via implementing the \textit{AnnotateCommunicationService} service interface, which can ensure compatibility between \name and external modules. The OneIE wrapper is one example of an external module.  %

\subsection{Mix and Match Modules: SM-KBP}

As another example application, we reconfigured our pipeline for the NIST SM-KBP 2020 Task 1 evaluation, which aims to produce document-level knowledge graphs.\footnote{\url{https://tac.nist.gov/2020/KBP/SM-KBP/index.html}} Each given document may be in English, Russian, or Spanish. On a development set consisting solely of text-only documents,\footnote{AIDA M36, \texttt{LDC2020E29}.} we started with initial predictions made by GAIA \cite{li-etal-2020-gaia}, for entity clusters, entity types, events and relations. Our goal was to recluster and relabel the a dataset for knowledge extraction. 

Our pipeline consisted of the multilingual coreference resolution (using the predetermined mention from GAIA) and hierarchical entity typing models discussed in this paper, followed by a separate state-of-the-art argument linking model \cite{chen-etal-2020-joint-modeling}. We found improved performance\footnote{This evaluation metric is specific to the NIST SM-KBP 2020 task. It takes entity types into account.} with entity coreference (from 29.1 F1 to 33.3 F1), especially in Russian (from 26.2 F1 to 33.3 F1), likely due to our use of multilingual data and contextualized encoders. The improved entity clusters also led to downstream improvements in entity typing and argument linking. This example highlights the ability to pick out subcomponents of \name and customize according to the downstream task.

\section{Usage}

We present two methods to interact with the pipeline. The first is a Docker container which contains the libraries, code, and trained models of our pipeline. This is intended to run on batches of documents. As a lighter demo of some of the system capabilities, we also have a web demo intended to interactively run on shorter documents.

\paragraph{Docker}

Our Docker image\footnote{\url{https://hub.docker.com/r/hltcoe/lome}} consists of the four core modules: FrameNet parser, coreference resolution, entity typing, and temporal resolution. Furthermore, there are two options for entity typing: a fine-grained hierarchical model (with the AIDA typing ontology) and a coarse-grained model (with the KAIROS typing ontology). The container and documentation is available on Docker Hub.

As some modules depend on GPU libraries, the image also requires NVIDIA-Docker support. Since there is a high start-up (time) cost for using Docker and loading models, we recommend using this container for batch processing of documents. Further instructions for running can be found on the \name Docker Hub page.

\paragraph{Web Demo}

We make a few changes for the web demo.\footnote{\url{https://nlp.jhu.edu/demos/lome/}} To reduce latency, we preload the models into memory and we do not write the \textsc{Concrete} communications to disk. At the cost of modularity, this makes the demo lightweight and fast, allowing us to run it on a single 16GB CPU-only server. To present the predictions, our front-end uses \texttt{AllenNLP-demo}.\footnote{\url{https://github.com/allenai/allennlp-demo}.} 

In addition, the web demo is currently limited to FrameNet parsing and coreference resolution, as other models will increase latency and may impede usability. The web demo is intended to highlight only some of the system's capabilities, like its ability to process multilingual documents. 

\section{Conclusions}

To facilitate increased interest in multilingual document-level knowledge extraction with large ontologies, we create and demonstrate \name, a system for event and entity knowledge graph creation. Given input text documents, \name runs a full FrameNet parser, coreference resolution, fine-grained entity typing, and temporal relation prediction. Furthermore, each component uses \textit{XLM-R}, allowing our system to support a broader set of languages than previous systems. The pipeline uses a standardized data schema, which invites extending the pipeline with additional modules. By releasing both a Docker image and presenting a lightweight web demo, we hope to enable the community to build on top of \name for even more comprehensive information extraction.

\section*{Acknowledgments}

We thank Anton Belyy, Kenton Murray, Manling Li, Varun Iyer, and Zhuowan Li for helpful discussions and feedback. This work was supported in part by DARPA AIDA (FA8750-18-2-0015) and KAIROS (FA8750-19-2-0034). The views and conclusions contained in this work are those of the authors and should not be interpreted as necessarily representing the official policies, either expressed or implied, or endorsements of DARPA or the U.S. Government. The U.S. Government is authorized to reproduce and distribute reprints for governmental purposes notwithstanding any copyright annotation therein.

\bibliography{anthology,eacl2021}
\bibliographystyle{acl_natbib}

\appendix

\end{document}